\documentclass{article}

\usepackage{arxiv}

\usepackage[utf8]{inputenc} 
\usepackage[T1]{fontenc}    
\usepackage{hyperref}       
\usepackage{url}            
\usepackage{booktabs}       
\usepackage{amsfonts}       
\usepackage{nicefrac}       
\usepackage{microtype}      
\usepackage{lipsum}		
\usepackage{graphicx}
\usepackage{natbib}
\usepackage{doi}
\usepackage{listings}

\usepackage{hyperref}

\usepackage{multirow}
\usepackage{caption}
\usepackage{subcaption}
\usepackage{float}
\usepackage{tabulary}
\usepackage{makecell}

\title{Scalable Knowledge Graph Construction and Inference on Human Genome Variants}

\author{ \href{https://orcid.org/0000-0001-9102-0709}{\includegraphics[scale=0.06]{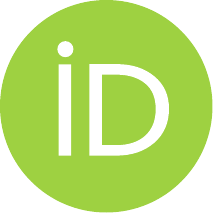}\hspace{1mm}Shivika Prasanna} \\
	University of Missouri-Columbia\\
	Columbia, MA 65201 \\
	\texttt{spn8y@umsystem.edu} \\
	\And
	\hspace{1mm}Deepthi Rao \\
	University of Missouri-Columbia\\
	Columbia, MA 65201 \\
	\texttt{raods@health.missouri.edu} \\
    \And
    \hspace{1mm}Eduardo Simoes \\
    University of Missouri-Columbia\\
	Columbia, MA 65201 \\
    \texttt{simoese@health.missouri.edu}
    \And 
    \hspace{1mm}Praveen Rao \\
    University of Missouri-Columbia\\
	Columbia, MA 65201 \\
    \texttt{praveen.rao@missouri.edu}
}

\renewcommand{\shorttitle}{\textit{arXiv} Template}

\hypersetup{
pdftitle={A template for the arxiv style},
pdfsubject={q-bio.NC, q-bio.QM},
pdfauthor={Shivika~Prasanna},
pdfkeywords={Knowledge graphs, Variant-level genomic information, RNA-seq human genome variants, Graph machine learning},
}

\begin{document}
\maketitle

\begin{abstract}
	Real-world knowledge can be represented as a graph consisting of entities and relationships between the entities. The need for efficient and scalable solutions arises when dealing with vast genomic data, like RNA-sequencing. Knowledge graphs offer a powerful approach for various tasks in such large-scale genomic data, such as analysis and inference. In this work, variant-level information extracted from the RNA-sequences of vaccine-naïve COVID-19 patients have been represented as a unified, large knowledge graph. Variant call format (VCF) files containing the variant-level information were annotated to include further information for each variant. The data records in the annotated files were then converted to Resource Description Framework (RDF) triples. Each VCF file obtained had an associated CADD scores file that contained the raw and Phred-scaled scores for each variant. An ontology was defined for the VCF and CADD scores files. Using this ontology and the extracted information, a large, scalable knowledge graph was created. Available graph storage was then leveraged to query and create datasets for further downstream tasks. We also present a case study using the knowledge graph and perform a classification task using graph machine learning. We also draw comparisons between different Graph Neural Networks (GNNs) for the case study.
\end{abstract}


\section{Introduction}
\label{introduction}
The Human Genome project \footnote{\url{https://www.genome.gov/human-genome-project}} aimed to sequence the entire human genome, resulting in an official gene map. The gene map has offered crucial insights into the human blueprint, accelerating the study of human biology and advancements in medical practices. This information has been represented in Variant Calling Format (VCF) files that store small-scale information or genetic variation data. 

Variants are genetic differences between healthy and diseased tissues or between individuals of a population. Analyzing variants can tremendously help prevent, diagnose, and even treat diseases. The process of analyzing these genetic differences or variations in DNA sequences and categorizing their functional significance is called variant analysis. RNA sequencing is similar to DNA sequencing but differs in its extraction. RNA is extracted from a sample and then reverse transcribed to produce what is known as copy or complementary DNA called cDNA. This cDNA is then fragmented and run through a next-gen sequencing system. Examining DNA provides a static picture of what a cell or an organism might do, but measuring RNA tells us precisely what the cell or organism is doing. Another advantage of RNA sequencing is that molecular features sometimes can only be observed at the RNA level.

Variant calling pipeline is the process of identifying variants from sequence data. To measure the deleteriousness of a variant, the Combined Annotation Dependent Depletion (CADD) \cite{rentzsch2019cadd,rentzsch2021cadd} scores tool is used. CADD evaluates or ranks the deleteriousness of a single nucleotide, insertion, and deletion variants in the human genome. The COVID-19 genetic data discussed in this paper was collected from the European Nucleotide Archive (ENA)\footnote{\url{https://www.ebi.ac.uk/ena/browser/view/SRR12095153}}\cite{ieeedataport}. 

The field of genomics has made remarkable strides in unraveling the genetic basis of numerous diseases, exploring evolutionary relationships, and understanding the molecular mechanisms underlying essential biological processes. Nevertheless, as genomic data continues to grow in size and complexity, researchers are faced with the daunting task of managing, integrating, and interpreting this wealth of information effectively. Traditional data storage and analysis methods often fall short when capturing the intricate relationships and context-dependent associations prevalent in genomic datasets.

Representing genomic data as knowledge graphs allows vast and diverse information from various sources to be integrated. These specialized graph structures, which model entities as nodes and relationships as edges, provide an ideal framework for integrating and organizing diverse biological information from multiple sources.  Furthermore, it allows for efficient querying and indexing and supports inference and new knowledge discovery. 

In this paper, we first introduce the previously explored work that combines the powers of knowledge graphs with vast genomic data. Next, we elaborate on the data collection pipeline and construction of our knowledge graph. We discuss the ontology and various components of our graph. Lastly, we demonstrate a use case for our graph using graph machine learning for a classification task. In this use case, we also draw comparisons between the different graph neural networks such as Graph Convolutional Network (GCN) and GraphSAGE.

\section{Related Work}
\label{sec:related-work}

The integration of knowledge graphs and genomic data remains relatively less explored, with limited prior research works. In this section, we delve into the relevant works that explore the interconnection of knowledge graphs with genomic data and deep learning in the domain of genetics. 

\cite{feng2023genomickb} presented a knowledge graph, GenomicKB, that integrates diverse types of genomic data and knowledge into a single, unified framework. The primary aim of the large graph was to provide a comprehensive and contextual representation of genomic information, enabling efficient data integration, analysis, and knowledge discovery. By consolidating various sources of genomic information, including data from genomic databases, experimental studies, literature, and public repositories, GenomicKB offers a comprehensive and holistic view of the genetic landscape.

\cite{reese2021kg} propose KG-COVID-19, a knowledge graph framework that was created by integrating heterogenous types of data available on SARS-CoV-2 and related viruses (SARS-CoV, MERS-CoV). KG-COVID-19 can be utilized for several downstream tasks such as machine learning, hypothesis-based querying, and browsable user interface to enable users to explore and discover new relationships in the data. The authors also utilized the proposed framework to create a knowledge graph for COVID-19 response.

\cite{harnoune2021bertclinical} proposed an innovative approach for constructing knowledge graphs from clinical data using BERT (Bidirectional Encoder Representations from Transformers) model. This work focused on creating biomedical knowledge graphs by leveraging BERT's contextual understanding capabilities. BERT was applied to process biomedical text data, including clinical records, scientific literature, and other relevant sources. By doing so, meaningful and contextually rich information was extracted from the text.

Deep learning has significantly influenced a wide range of domains, with genomic studies being particularly impacted. \cite{liu2020deepcdr} introduced a new method called DeepCDR that uses deep learning methods for predicting the response of cancer cells to different drugs. The primary objective of DeepCDR was to facilitate effective cancer treatment by accurately predicting how specific cancer cells can respond to different drugs.

\cite{lanchantin2019graph} presented ChromeGCN for predicting epigenetic states using sequences and 3D genome data. ChromeGCN leverages graph convolutional networks (GCNs) to predict the epigenetic states of the genomic regions. The genomes have been represented as a graph where the genomic regions are the nodes and relationships between the genomes have been represented as the edges. The predictive power of ChromeGCN has significant implications as it enables the identification of functional genomic elements and regulatory regions, providing insights on the mechanisms that govern gene regulation and cellular function.

\cite{sun2018knowledge} proposed a new method called KGBSVM (Kernelized Generalized Bayesian Rule Mining (KGBRM) with Support Vector Machines (SVM)) that analyzes high-dimensional genome data that can improve classification accuracy on general classification tasks that can be binary or multi-class. The KGBSVM method aims to enhance the accuracy and efficiency of these types of classification tasks.





\section{Knowledge Graph}
\label{sec:knowledge-graph}
In this section, the construction of knowledge graphs is explained. First, we describe how the data was collected and the workflow for the process. Then, we highlight the process of annotating the data, followed by the ontology description. Lastly, we discuss the representation of VCF files and CADD scores in a knowledge graph.

\subsection{Data Collection}
\begin{figure*}[tbh]
    \centering
    \includegraphics[width=0.8\textwidth]{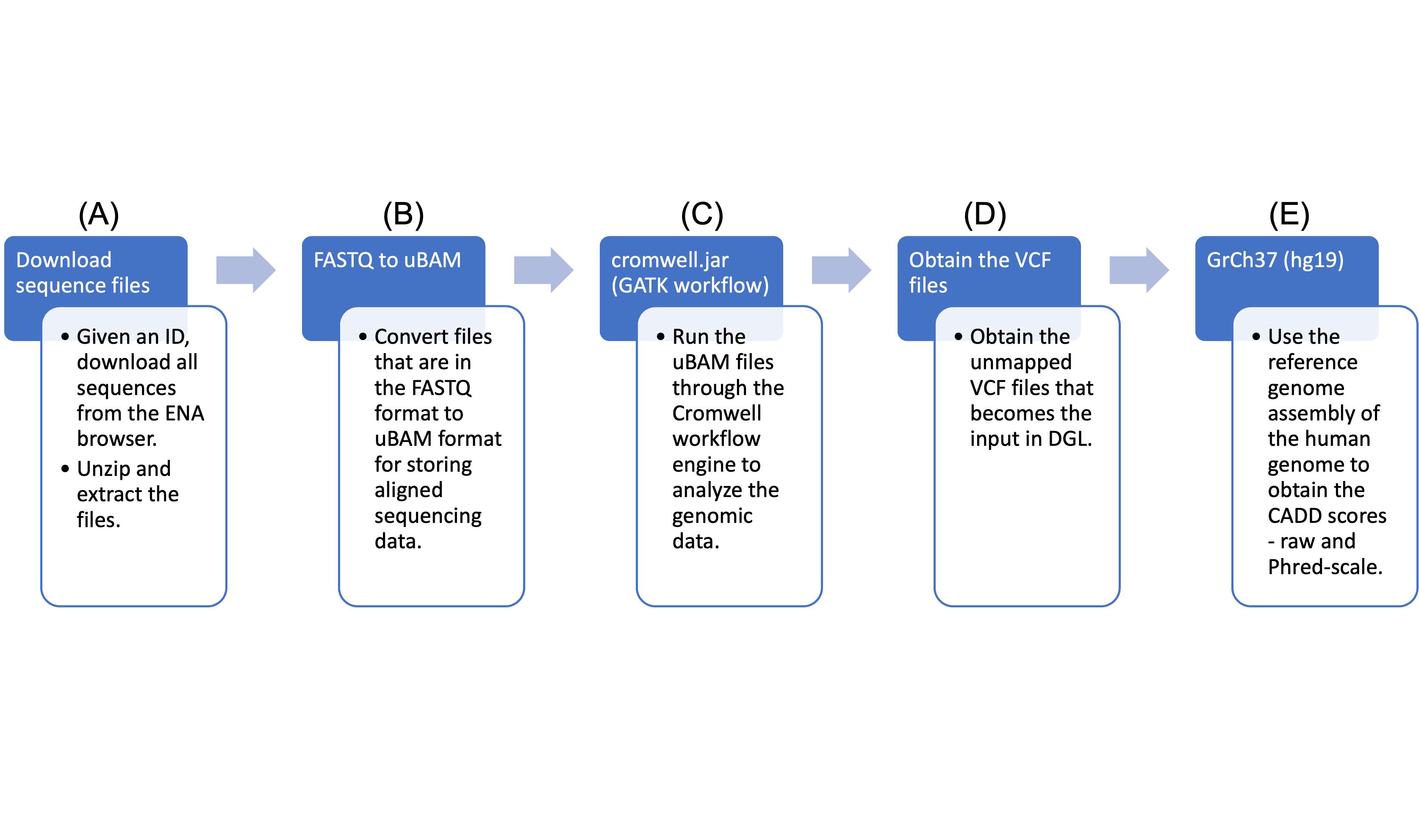}
    \caption{Workflow to show how the raw data was collected to be further processed and collated into a dataset.}
    \label{fig:seq-workflow}
\end{figure*}

COVID-19 RNA sequence IDs were first collected from the European Nucleotide Archive. A total of 3,716 VCF files were collected till March 2023. The workflow has been shown in Figure \ref{fig:seq-workflow}.

\begin{itemize}
    \item FASTQ files (part A): These IDs were utilized to download the RNA sequences, which were in FASTQ \cite{li2008mapping,li2009fast} format. FASTQ is a text-based file format used for storing both a biological sequence (usually nucleotide sequence) and its corresponding quality scores. It is commonly used to represent the output of high-throughput sequencing technologies. The FASTQ file format consists of a series of records, each of which contains four lines of text: the first line starting with `@' contains a sequence identifier, the second line contains the actual nucleotide sequence, the third line starts with `+' and may optionally contain additional information about the sequence, and the fourth line contains quality scores encoded as ASCII 10 characters. The quality scores indicate the confidence in the accuracy of each base call and are typically represented as Phred scores.
    \item uBAM files (part B): The FASTQ files were then converted to unmapped BAM (uBAM) \footnote{\url{https://gatk.broadinstitute.org/hc/en-us/articles/360035532132-uBAM-Unmapped-BAM-Format}} files for storing aligned sequencing data. It is an uncompressed version of a BAM file, which is a binary representation of sequence alignment data. A uBAM file contains unmapped reads, meaning reads that could not be confidently aligned to a reference genome. These reads can be used for downstream analysis, such as de novo assembly, quality control, and identification of novel sequences. Unlike a BAM file, a uBAM file does not contain alignment information and is, therefore smaller in size, making it more convenient for storage and transfer.
    \item GATK workflow (part C): The uBAM files were passed through the Genomic Analysis Toolkit (GATK) workflow \cite{mckenna2010genome} that converts the files into Variant Calling Format (VCF) \cite{danecek2011variant} files. It is a comprehensive toolkit developed by the Broad Institute that includes various tools and algorithms for processing genomic data, such as read mapping, local realignment, base quality score recalibration, variant calling, and variant filtering. The workflow involves several steps, including preprocessing of raw data, alignment to a reference genome, post-alignment processing, and variant discovery and annotation.
    \item The unannotated VCF files that were obtained as the result of the workflow have been shown in part D. For each VCF file, there is also a corresponding CADD scores file that was obtained using GrCh37 \footnote{\url{https://krishna.gs.washington.edu/download/CADD/v1.6/GRCh37}} through the workflow, as shown in part E.
\end{itemize}

\subsection{Data Annotations}
\label{data-annotations}
Once the RNA sequencing was completed, two main files were generated for each RNA-seq ID – a VCF file and a CADD scores file in TSV format.

For further annotations, SnpEff \cite{cingolani2012program}, a command-line, variant annotation, and effect prediction tool, was utilized. This tool annotates and predicts the effects of genetic variants. SnpEff takes as input the predicted variants (SNPs, insertions, deletion, and MNPs) and produces a file with annotations of the variants and the effects they produce on known genes. 

\begin{figure*}[tbh]
    \centering
    \includegraphics[keepaspectratio, width=\textwidth]{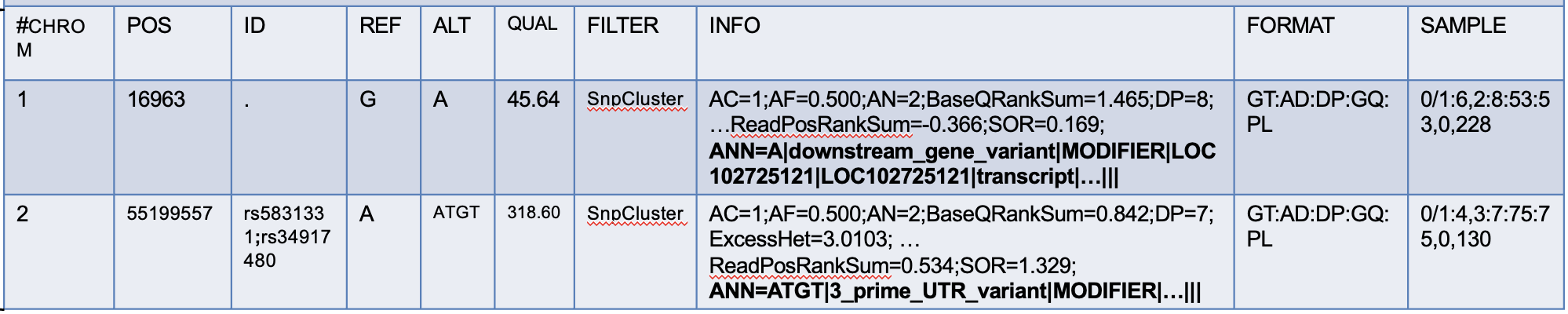}
    \caption{Additional annotations by the SnpEff tool.}
    \label{fig:ann-field}
\end{figure*}

SnpEff classifies variants as single nucleotide polymorphisms (SNPs), insertions, deletions, multiple-nucleotide polymorphisms or an InDel. While the original VCF file contains the INFO field, SnpEff adds additional annotations to this field to further describe the variation. In the process, it also updates the header fields. This field is tagged by `ANN', which is pipe symbol separated and provides a summary of the predicted effects of a genetic variant on each affected transcript. Figure \ref{fig:ann-field} shows the ANN field highlighted in bold.

A variant may have one or more annotations and multiple annotations are comma-separated. There are several fields within the ANN tag, mainly:

\begin{itemize}
    \item Allele (ALT): Information on alternate alleles that are predicted to cause a functional impact on a gene or protein
    \item Annotation (effect): Type of effect caused by the variant on the transcript
    \item Putative impact: Qualitative measure of the impact of the variant on the transcript
    \item Gene name: Name of the affected gene
    \item Gene ID: Unique identifier of the affected gene
\end{itemize}

\subsection*{Ontology}

\begin{figure*}[tbh]
    \centering
    \includegraphics[keepaspectratio, width=\textwidth]{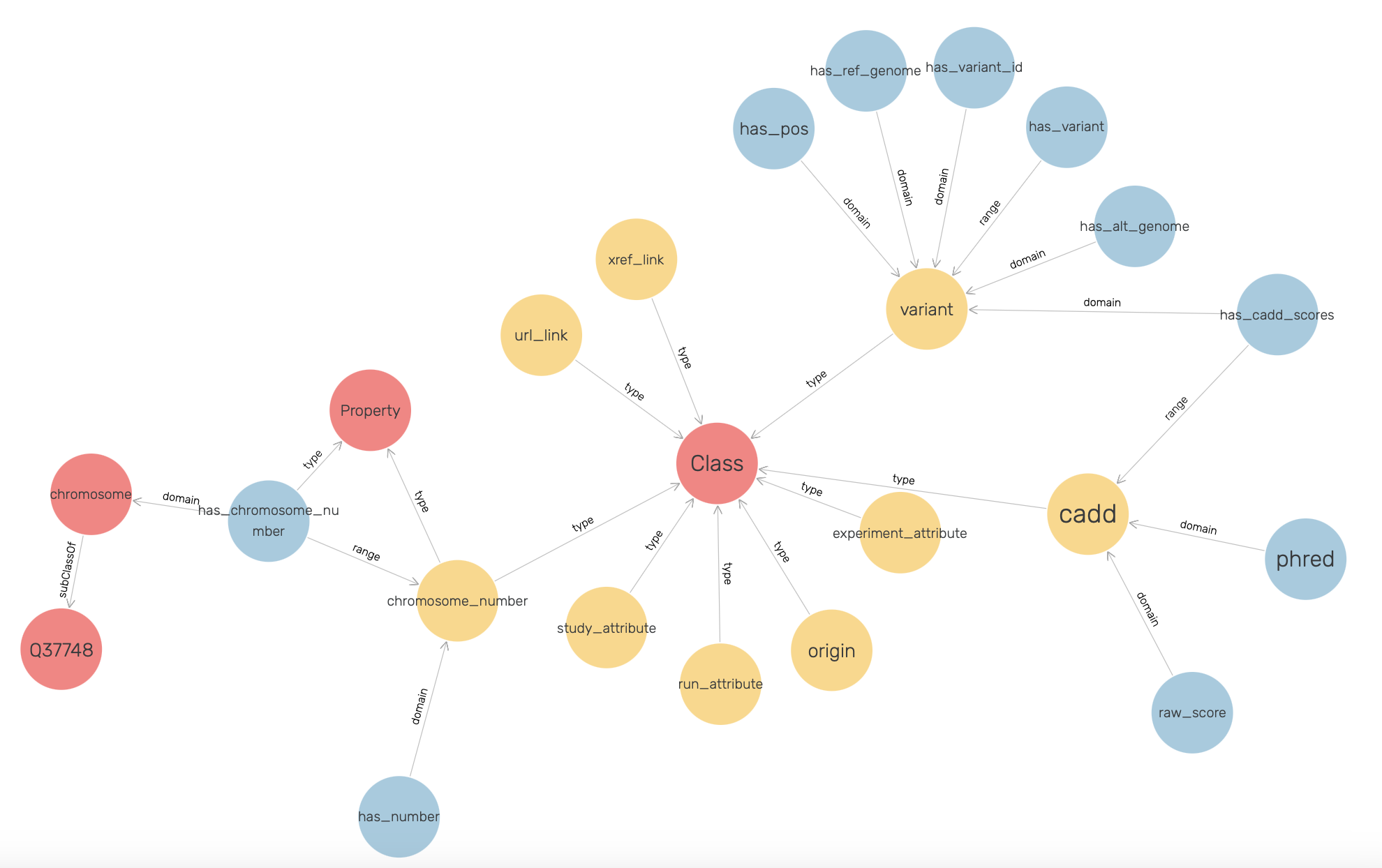}
    \caption{Ontology of the knowledge graph.}
    \label{fig:ontology}
\end{figure*} 

A knowledge graph is represented using an ontology, where the ontology can be represented using a formal language such as RDF (Resource Description Framework) or OWL (Web Ontology Language), or another domain-specific language. The ontology in this work has been represented using RDF. Each node-edge-node is represented as a triple by RDF. In a triple, the subject defines the first node, object defines the second node. The predicate defines the edge or relation joining the two nodes. A triple always ends with a dot. 

An ontology mainly consists of classes, properties, and relationships. In Figure \ref{fig:ontology}, the sub-classes are depicted by red nodes, the classes are by yellow nodes, and the relations by blue nodes. The description of the classes has been given in Table \ref{tab:classes-description} and the description of the properties has been given in Table \ref{tab:properties-description}.

\begin{table}[t!]
\centering
\caption{Description of the classes in the ontology.}
\label{tab:classes-description}
\begin{tabular}{|lp{30em}|}
\hline
CLASS & DEFINITION  \\
\hline
Chromosome Number & Identifier of the chromosome; values can be `1', `2', …, `22', `X', `Y', `MT' \\
Origin & Unique identifier of the variant annotated by SPARQLing Genomics tool \\
Variant & Encapsulates the different types of genomic alterations that can occur \\
CADD & Encapsulates the different types of scores that can occur \\
xref\_link & Type of annotation that provides a link between different resources or databases \\
url\_link & Access link to experiment label \\
study\_attribute & Metadata that describes the experimental design, data processing, and other aspects of a sequencing study \\
run\_attribute & Metadata that describes the sequencing run \\
experiment\_attribute & Metadata that describes the overall experimental design and goal of the experiment \\
\hline

\end{tabular}
\end{table}

\begin{table}[t!]
\centering
\caption{Description of the properties in the ontology.}
\label{tab:properties-description}
\begin{tabular}{|lp{15em}p{6.5em}p{5em}|}
\hline
PROPERTY & DEFINITION & DOMAIN & RANGE \\
\hline
has\_pos & Variant position & Variant & Integer \\
has\_ref\_genome & Reference genome at that position & Variant & String \\
has\_alt\_genome & Alternate genome at that position & Variant & String \\
has\_variant\_id & Unique identifier of the variant & Variant & String \\
has\_variant & Unique name given to the variant & Variant & String \\
has\_cadd\_scores & Variant has associated CADD scores & Variant & CADD \\
has\_chromosome\_number & Chromosome has a chromosome number & Chromosome & String \\
phred & Phred-scaled score & CADD & Long \\
raw\_score & Raw CADD score & CADD & Long \\
\hline

\end{tabular}
\end{table}

In the defined ontology, chromosome and variant are both domain classes and a chromosome has an associated chromosome number to be able to connect all similar chromosomes as an extension, and a variant has an associated variant ID. A variant has a reference and alternate genome. 

The ontology also explicitly defines CADD as a class where a variant has CADD scores represented by both raw score and Phred-scale score, as properties of the CADD class. The ontology description has been given in Table \ref{tab:ontology-description}.

\begin{table}[t!]
\centering
\caption{Domain, properties, and ranges for the ontology.}
\label{tab:ontology-description}
\begin{tabular}{|lp{9em}p{10em}p{7em}|}
\hline
ENTITY & RDF:PROPERTY & DOMAIN & RANGE \\
\hline
Chromosome & Type & N/A & Wiki:Q37748 \\
 & SubClassOf & N/A & Wiki:Q37748 \\
\hline
has\_chromosome\_number & Type & N/A & Property \\
 & Domain & Chromosome & N/A \\
 & Range & chromosome\_number & N/A \\
\hline
chromosome\_number & Type & N/A & Class \\
\hline
has\_number & Type & N/A & Property \\
 & Domain & chromosome\_number & N/A \\ 
 & Range & xsd:int & N/A \\ 
\hline
Variant & Type & N/A & Class \\
\hline
has\_variant & Type & N/A & Property \\ 
 & Domain & Chromosome & Variant \\ 
 & Range & Variant & N/A \\
\hline
has\_pos & Type & N/A & Property \\ 
 & Domain & Variant & xsd:string \\ 
 & Range & xsd:int & N/A \\
\hline
has\_ref\_genome & Type & N/A & Property \\
 & Domain & Variant & xsd:string \\
 & Range & xsd:string & N/A \\
\hline
has\_alt\_genome & Type & N/A & Property \\
 & Domain & Variant & xsd:string \\
 & Range & xsd:string & N/A \\
\hline
CADD & Type & N/A & Class \\
\hline
has\_cadd\_score & Type & N/A & Property \\
 & Domain & Variant & CADD \\
 & Range & CADD & N/A \\
\hline
raw\_score & Type & N/A & Property \\
 & Domain & CADD & xsd:long \\
 & Range & xsd:long & N/A \\
\hline
phred & Type & N/A & Property \\
 & Domain & CADD & xsd:long \\
 & Range & xsd:long & N/A \\
\hline

\end{tabular}
\end{table}

\subsection{Conversion of VCF files to Knowledge Graphs}
\label{vcf-to-kg}

To transform the data in VCF, SPARQLing Genomics \footnote{\url{https://gitlab.com/roelj/sparqling-genomics}} was utilized. SPARQLing Genomics is an open-source platform for querying and analyzing genomic data using the Semantic Web and Linked Data technologies. The platform provides an easy-to-use interface as well, that has been built to support SPARQL queries and various SPARQL query features, including sub-queries, filters, and aggregates. SPARQLing Genomics provides several in-built, ready-to-use tools, one of which is vcf2rdf that converts VCF data into RDF triples. 

The triples generated by the tool consist of uniquely identifiable names having symbolic value and literal values like numbers or text. 
\newpage
\begin{lstlisting}
#CHROM POS   ID REF ALT QUAL    FILTER    INFO sample
   1  16963  .   G   A  45.64 SnpCluster AC=1;AF=0.500;AN=2;BaseQRankSum=1.465;DP-8;ExcessHet=3.0103;FS=0.000;MLEAC=1;MLEAF=0.500;MQ=60.00;MQRankSum=0.000;QD=5.70;ReadPosRankSum=-0.366;SOR=0.169GT:AD:D:GQ:PL0/1:6,2:8:53:53,0,228
\end{lstlisting}

The following is an example of how a variant position (example shown above) is translated into a triple by the tool.

\begin{lstlisting}
<origin://4a37140cdc877d90ffe258a8151f27e@0> <http://biohackathon.org/resource/faldo#position> "16963"^^<http://www.w3.org/2001/XMLSchema#integer> .
\end{lstlisting}





As seen in the above example, the variant position is described with Feature Annotation Location Description Ontology (FALDO) \cite{bolleman2016faldo}. For other features not defined by FALDO, the URI is customized to the tool.

Each VCF file eventually corresponds to one large knowledge graph that was originally stored in an N3 format. N3 format is one of the several formats supported by RDF and can be considered as a shorthand non-XML serialization of RDF models. However, to accommodate the accessionID that would map to an unidentified patient, the N3 serialization was converted to NQ format with the accessionID as the named graph. An example of a triple from an N3 file has been given here.

\begin{lstlisting}
<origin://4a37140cdc877d90ffe2b58a8151f27e@0> <sg://0.99.11/vcf2rdf/variant/REF> <sg://0.99.11/vcf2rdf/sequence/G>  .
\end{lstlisting}

The triple was then converted to NQ format which yielded the following triple:

\begin{lstlisting}
    <origin://4a37140cdc877d90ffe2b58a8151f27e@0>   <sg://0.99.11/vcf2rdf/variant/REF> <sg://0.99.11/vcf2rdf/sequence/G> <sg://SRR13112995>.
\end{lstlisting}

The ontology was extended to accommodate the new relations generated by the tool. 

\subsection{Conversion of CADD score files to Knowledge Graphs}

The SnpEff and vcf2rdf tools were useful for converting VCF files to triples. However, CADD scores that were obtained through the pipeline were in tab-separated (TSV) format. To enrich the knowledge graphs, the CADD scores had to be translated to RDF triples as well. Therefore, the ontology for CADD scores was explicitly defined. 

\begin{figure}[!h]
    \centering
    \includegraphics[keepaspectratio, width=0.7\textwidth]{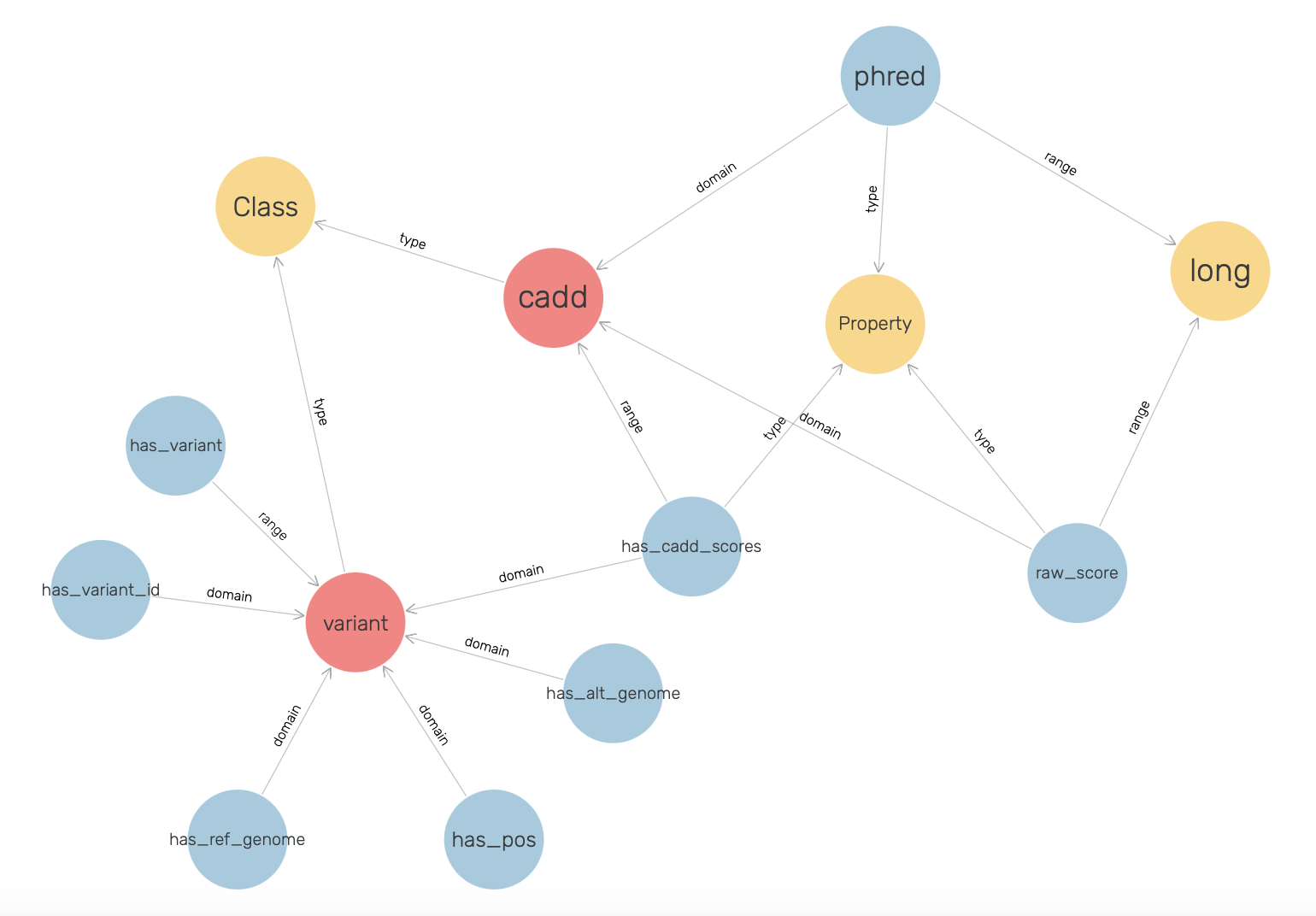}
    \caption{Ontology for CADD scores.}
    \label{fig:cadd-scores-ontology}
\end{figure}

To visualize the graph, GraphDB \footnote{\url{https://graphdb.ontotext.com/documentation/10.0/index.html}} has been utilized, and the ontology that was defined for CADD scores has been shown in Figure \ref{fig:cadd-scores-ontology}

These scores have been represented with respect to the described fields in the VCF files such as the chromosome, position, reference genome, and alternate genome. The raw scores and Phred-scale scores were obtained from the original TSV files.

The following is an example of a record in a TSV file for which the raw and phred scores map to chromosome 1 with position 16963, reference genome `G' and alternate genome `A' in the VCF file:

\begin{table}[H]
\centering
\label{tab:ex-cadd}
\begin{tabular}{|lp{5em}p{5em}p{5em}p{5em}p{5em}|}
\hline
\#Chrom & Pos & Ref & Alt & RawScore & PHRED \\
\hline
1 & 16963 & G & A & 0.900784 & 12.72 \\
\hline
\end{tabular}
\end{table}

Each data record, like the above example, was converted to a Turtle triple (TTL), another format supported by RDF. A TTL format writes a graph in a compact textual form. There are only 3 parts to this triple – subject, predicate, and object. An example of the above data record converted to a TTL triple is given below:

\begin{lstlisting}
   <http://sg.org/SRR13112995/1/variant1> a ns1:variant ;
        ns1:has_alt_genome "A" ;
        ns1:has_cadd_scores <http://sg.org/SRR13112995/1/variant1/cadd> ;
        ns1:has_pos 16963 ;
        ns1:has_ref_genome "G" .
\end{lstlisting}

\section{Results}

Each variant file represented a single knowledge graph, so to unify several knowledge graphs into one single large graph, BlazeGraph Database \footnote{\url{https://blazegraph.com/}} has been leveraged. This large knowledge graph was then queried to create a dataset for a case study. It is important to note that the edges are homogeneous in nature.

BlazeGraph is a high-performance, horizontally scalable, and open-source graph database that can be used to store and manage large-scale graph data. It has been designed to provide efficient graph querying and supports RDF data model that allows it to store and process both structured and semi-structured data. BlazeGraph uses a distributed architecture that can be easily integrated with other big data tools, such as Hadoop and Spark, to perform complex analytics on large-scale graph data.

BlazeGraph has been leveraged for efficiently querying the knowledge graphs to generate the dataset for Graph Neural Network downstream tasks. Other tools such as RIQ~\cite{RIQ2017,RIQ2016,RIQ2015} can be used to index and query RDF named graphs.

\begin{figure}[htb]
    \centering
    \includegraphics[keepaspectratio, width=0.95\textwidth]{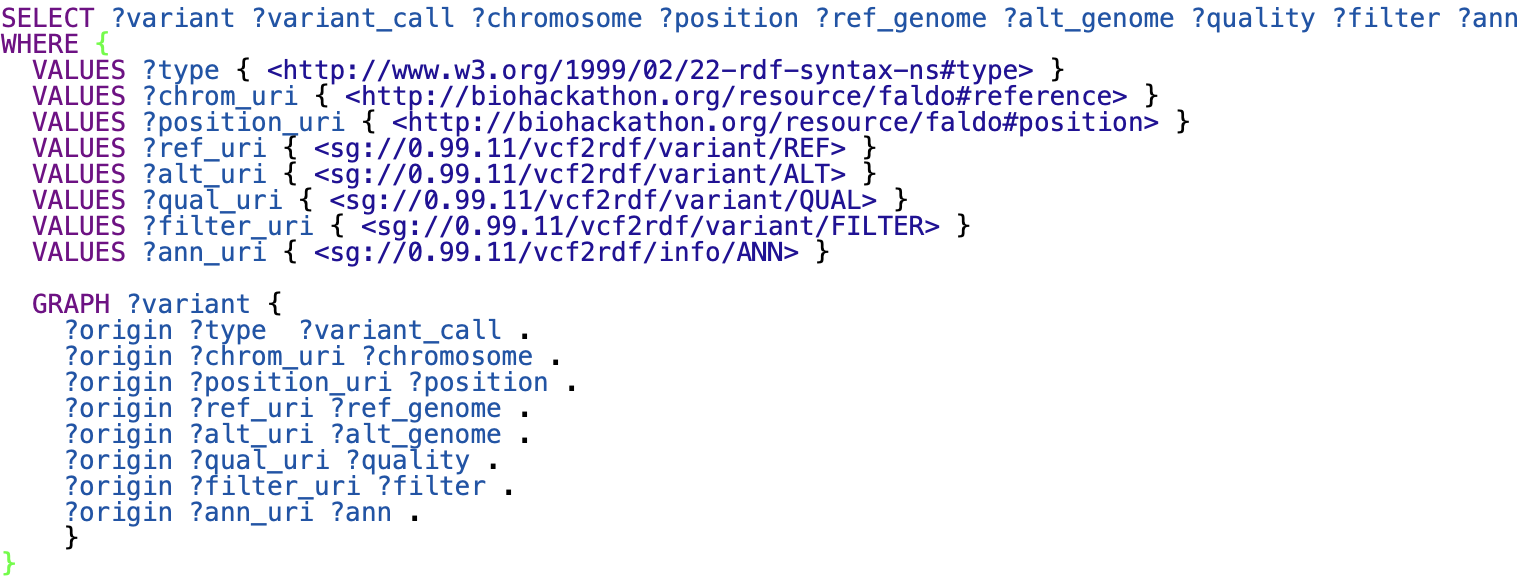}
    \caption{SPARQL query to extract variant and its properties.}
    \label{fig:simple-blazegraph-query}
\end{figure}

The total number of triples in the knowledge graph, after aggregating only 511 VCF files on a single machine, is as large as 3.1 Billion.

To query the variant that has properties such as chromosome, position, reference, and alternate genomes and the ANN field that had been annotated by the SnpEff tool, the SPARQL query in Figure \ref{fig:simple-blazegraph-query} was used.

Each chromosome has a position that has an associated reference genome, alternate genome, quality, and filter. 

\section{KG INFERENCE: CASE STUDY}

In this section, we will discuss a case study using the created knowledge graph. We aim to use the KG for a classification task using graph machine learning. The objective of the task is to classify a variant (or a node) into a CADD score category, given information about the variant. These categories allow researchers to identify where the actual raw score may lie for the variant. This would allow researchers to focus on the more likely deleterious variants first, depending on the nature of the problem.

For the classification task, we are leveraging the open-source graph-based library called Deep Graph Library \cite{wang2019dgl}.

\subsection{Deep Graph Library}
To implement the task, Deep Graph Library (DGL), an open-source library supporting graph-based deep learning was utilized. DGL provides a set of high-level APIs for building scalable and efficient graph neural network models. With DGL, we can create, manipulate, and learn from large-scale graphs with billions of nodes and edges.

There are three main tasks supported by DGL:
\begin{itemize}
    \item Node Classification: Predict the class or label of a node in a graph based on its features.
    \item Link Prediction: Predict if there is a connection or an edge between two nodes.
    \item Graph Classification: Classify an entire graph into one or more classes or categories.
\end{itemize}

DGL represents a graph as a DGLGraph object, which is a framework-specific graph object. It requires the number of nodes and a list of source and destination nodes, where nodes and edges must have consecutive IDs starting from 0. Since DGL only accepts numeric input, all strings such as URI were mapped to integers. In this case study, node classification was used to classify variants into CADD score categories based on their features.

\subsection{Subset Graph Transformation}
The knowledge graph created needs to be transformed into the input required by DGL due to its design specifications. The transformation of the input required for graph machine learning needs to be done in a scalable and efficient way. 

\begin{figure}[!h]
    \centering
    \includegraphics[keepaspectratio, width=0.9\textwidth]{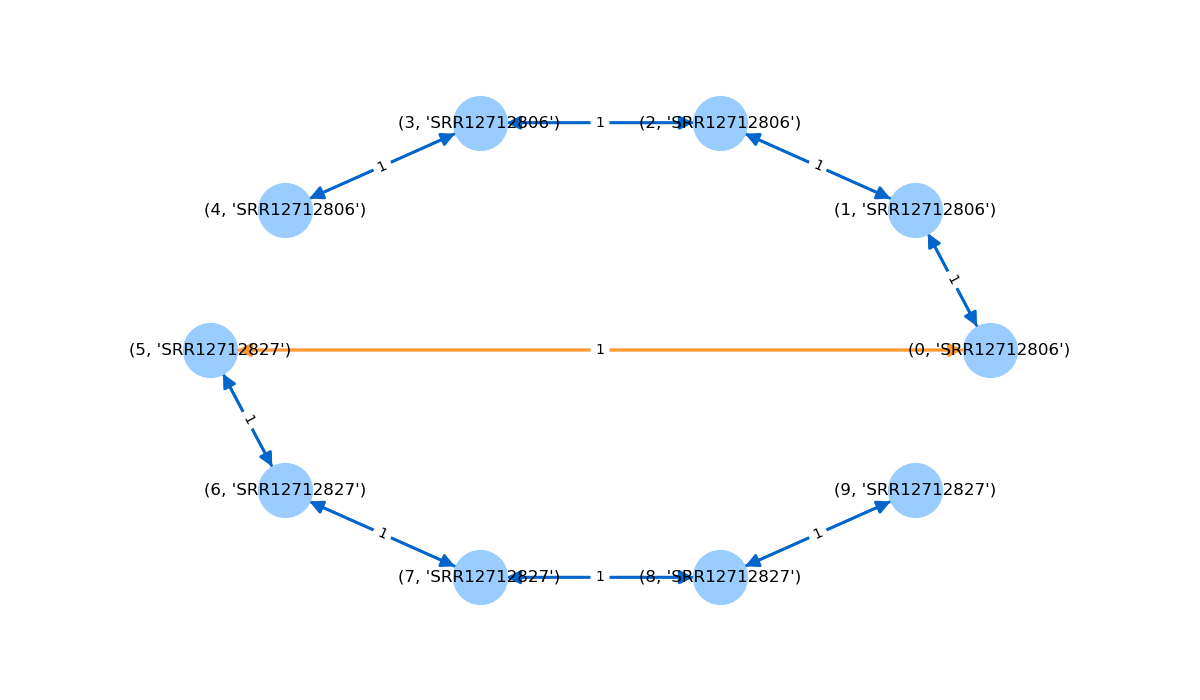}
    \caption{Graph where every raw variant ID is connected, within the same accession ID, as depicted by blue bidirectional arrows. Orange bidirectional arrow depicts two nodes connected across accession ID, by a shared variant ID.}
    \label{fig:graph-4}
\end{figure}

The graph was transformed in a way that all nodes belonging to the same accession ID or patient shared an edge and all nodes that share a variant ID also shared an edge. With this, no two nodes within the same accession ID can share a variant ID as this is unique to the accession ID. Therefore, nodes that share a variant ID are connected across accession IDs. This has been shown in Figure \ref{fig:graph-4}. 

\begin{figure}[!h]
    \centering
    \includegraphics[keepaspectratio, width=0.8\textwidth]{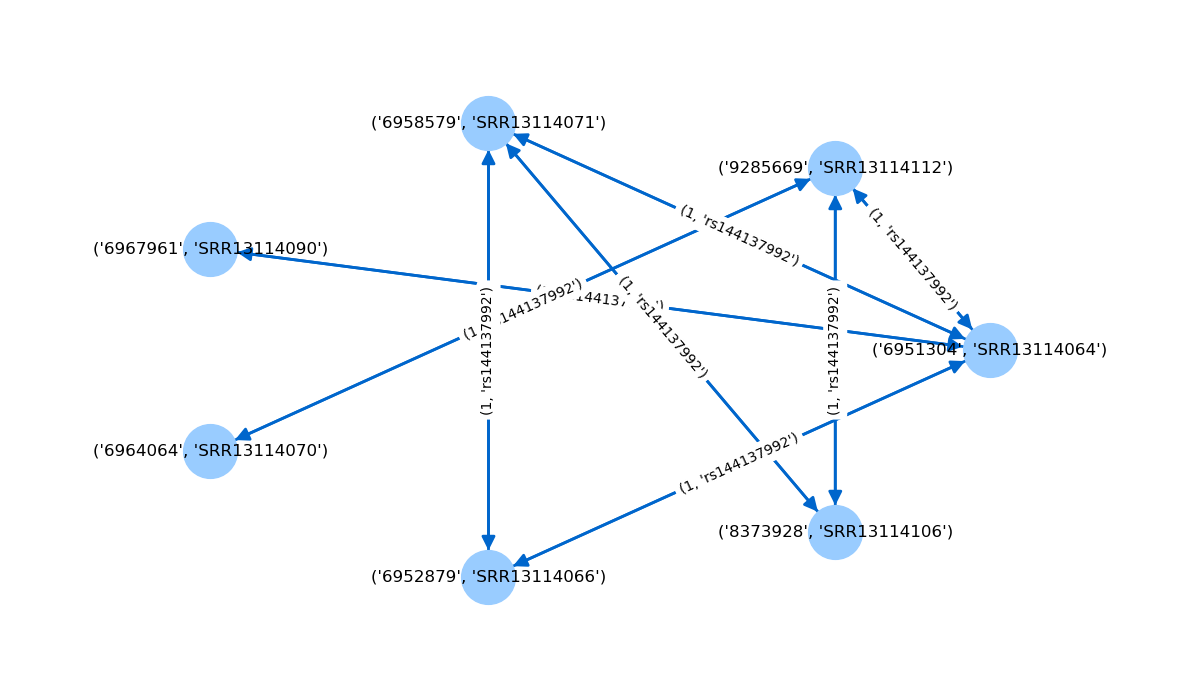}
    \caption{Projection of a denser graph where two nodes are connected if they share a variant ID.}
    \label{fig:graph-1}
\end{figure}

\begin{figure}[!h]
    \centering
    \includegraphics[keepaspectratio, width=0.8\textwidth]{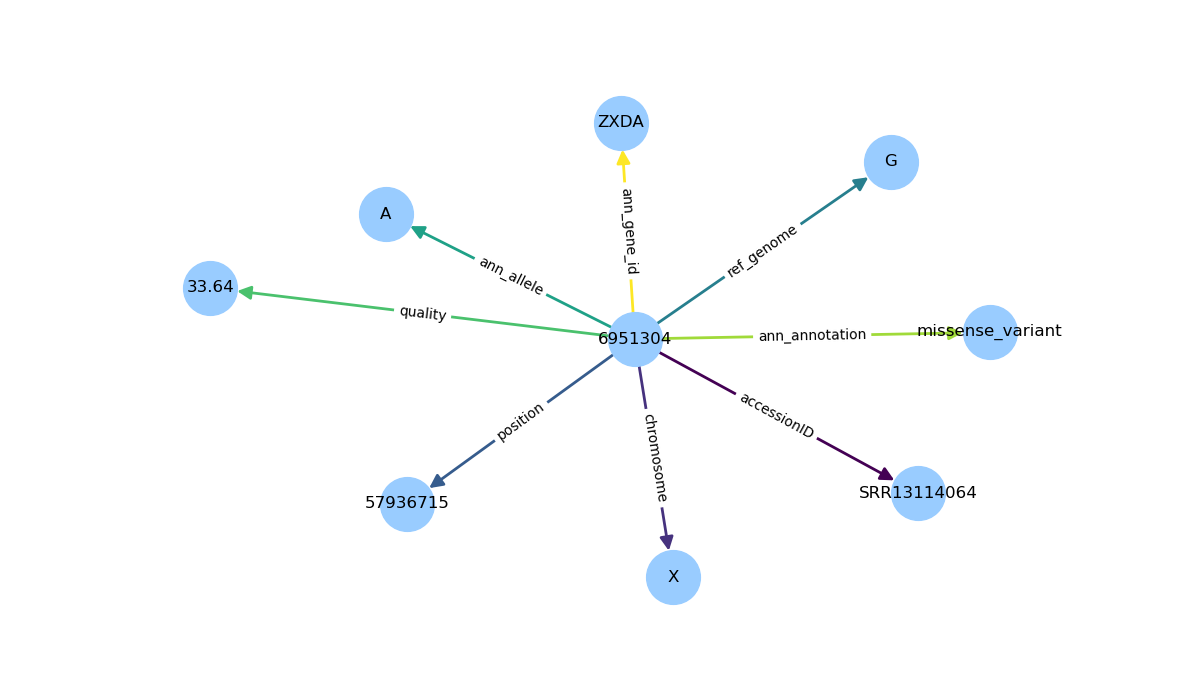}
    \caption{Node features of a sample node.}
    \label{fig:node-features}
\end{figure}

For the task, a portion of the graph was extracted as a projection where SPARQL queries were used to prepare the subset of the graph dataset that now contains variants connected to each other by a variant ID. That is, two nodes share an edge between them if they belong to the same variant ID. This projection has been shown in Figure \ref{fig:graph-1} and the node features for a sample node have been shown in Figure \ref{fig:node-features}.

As in Figure \ref{fig:node-features}, the accession ID, chromosome, position, reference genome, alternate genome, quality, and ANN information such as allele, annotation, gene name, and gene ID were considered as the node features.

The subset graph contained 190 unique accession IDs, with 73,661,245 nodes and 112,707,621 edges.

\subsection{Node Classification Task}
For this task, Graph Convolutional Network (GCN) \cite{zhang2019graph} and GraphSAGE \cite{hamilton2017sage} have been used. For both models, each node is associated with a feature vector. 

GCNs use node embeddings and adjacency matrices to compute new embeddings while training. Similar to CNN, the model weights and biases are first initialized to 1, and then a section of the graph is passed through the model. A non-linear activation function is used to compute predicted node embeddings for each node. Cross entropy loss is calculated to quantify the difference between the predicted node embeddings and the ground truth. Loss gradients are then computed to update the model using the Adam optimizer\cite{kingma2014adam} for this task. These steps are repeated until convergence.

GraphSAGE uses SAGEConv layers where for every iteration, the output of the model involves finding new node representation for every node in the graph. Mean is used as the aggregation function and ReLU activation function has been utilized. Adam optimizer was used for this model as well. One of the most noted properties of GraphSAGE is its ability to aggregate neighbor node embeddings for a given target node. Through the experiments conducted, this property was observed. GraphSAGE also generalizes better to unseen nodes because of its ability to perform inductive learning on graphs. 

The CADD (raw) scores have been categorized into 5 categories of equal distribution, primarily based on their frequencies generating target labels from 0 to 5. These categories have been shown in Table \ref{tab:cadd-categories}. The categories or bins for the target labels are based on the pattern found in the raw data used in this work. The categories are not arbitrary because there are no pre-defined CADD score categories. 

\begin{table}[!h]
\centering
\caption{CADD score categories for the classification task.}
\label{tab:cadd-categories}
\begin{tabular}{|l|p{7em}|}
\hline
RAW SCORES RANGE & CATEGORY \\
\hline
Less than 0 & 0 \\
0-1 & 1 \\
1-5 & 2 \\
5-10 & 3 \\
10-100 & 4 \\
\hline
\end{tabular}
\end{table}

\begin{figure}[!h]
    \centering
    \includegraphics[keepaspectratio, width=\textwidth]{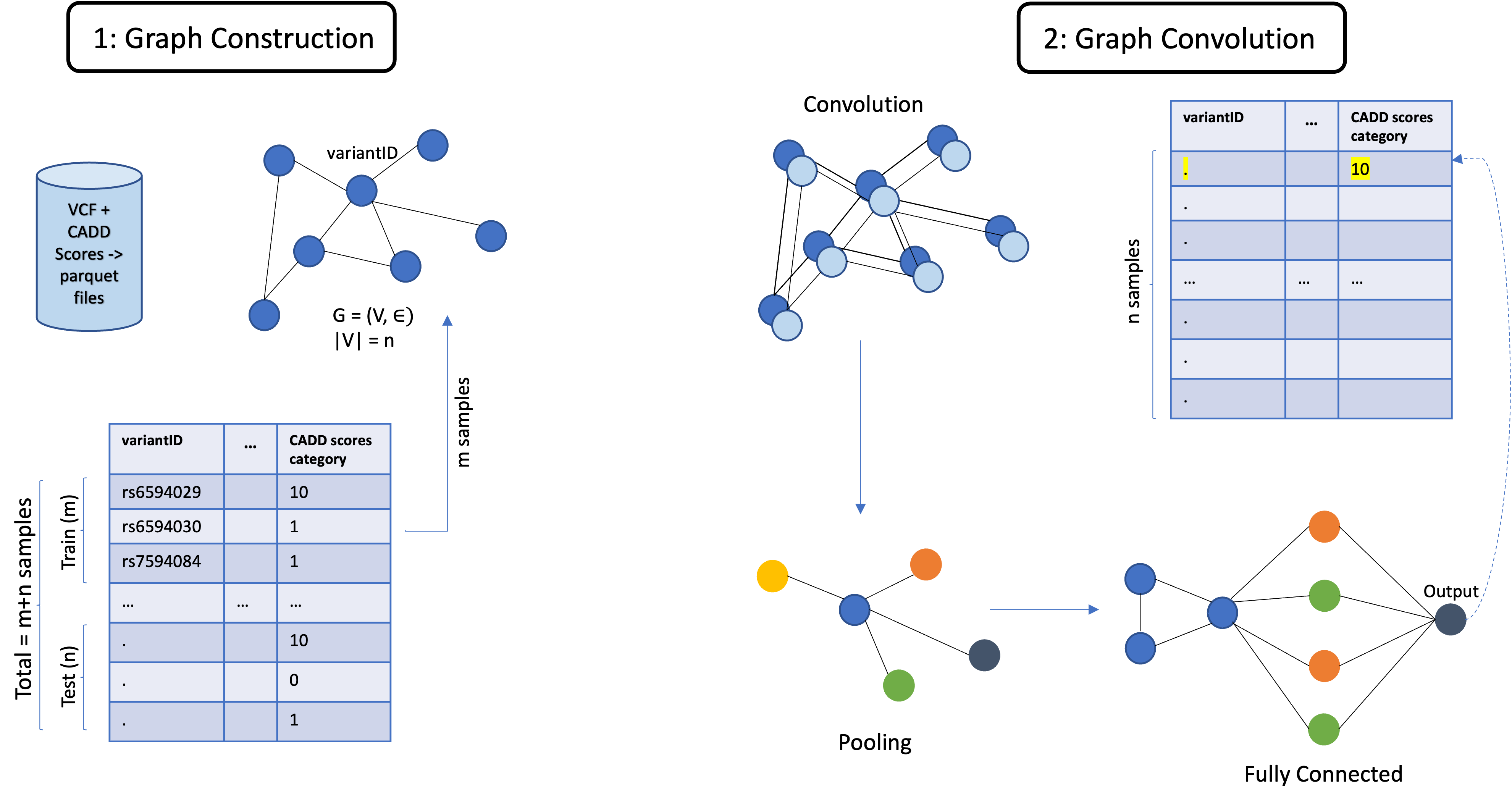}
    \caption{Architecure of GCN.}
    \label{fig:graph-conv-net}
\end{figure}

The architecture of GCN has been shown in Figure \ref{fig:graph-conv-net}. The architecture of GraphSAGE differs in the property of message passing between the nodes. This was crucial as the nodes in the input graph relied on several pieces of information from their neighboring nodes.

\subsection{Experiments \& Results}
\subsubsection{Experiment Setup}
The experiments were run on CloudLab \footnote{\url{https://www.cloudlab.us}}, a testbed for cloud computing research and new applications. CloudLab cluster at Clemson University was built in partnership with Dell. Clemson machines were used to carry out the standalone training experiments. They were carried out on nodes with 16 cores per node (2 CPUs), 12x4 TB disk drives in each node, plus 8x1 TB disks in each node. The nodes are configured with 256 GB of memory and 73TB of RAM. The operating system installed across all of them was Ubuntu 18.04.

Wisconsin machines were also used to carry out the standalone training experiments described in section 6.2. The CloudLab cluster at the University of Wisconsin was built in partnership with Cisco, Seagate, and HP. The cluster has 523 servers with a total of 10,060 cores and 1,396 TB of storage, including SSDs on every node. 

The nodes were chosen upon availability for the standalone node classification tasks.

\subsubsection{Experiment Results}
The data input was a graph with 73,661,245 nodes or vertices and 112,707,621 edges from 190 variants. The data was split into 60:20:20 (train:val:test) ratio and  Graph Convolutional Network (GCN) and GraphSAGE models were used to train on this dataset in separate experiments. Adam optimizer was employed along with Cross Entropy loss in both models. Grid search was conducted to find the most optimum hyperparameters.

Table \ref{tab:hyperparam-results-001} shows the hyperparameter tuning results for GraphSAGE and GCN models. For GraphSAGE, hyperparameter tuning was conducted keeping the learning rate of 0.001 constant over 1500 epochs with variable number of hidden layers. The hidden layers were chosen from 2, 8, 16, and 32 and it was observed that the highest validation accuracy of 85.8\% was observed when the model had 32 hidden layers. Similar experiments were performed on the GCN model, and it was observed that this model, too, yielded the highest validation accuracy of 84\% for 32 hidden layers. The best models in both sets of experiments were then evaluated on the test set, giving 86.67\% test accuracy by GraphSAGE and 82.6\% test accuracy by GCN. 

\begin{table}[H]
\centering
\caption{Performance comparison of GraphSAGE and GCN. The test accuracy is only computed for models with the best validation set accuracy. HL = Hidden Layers; LR = Learning Rate; VAL = Validation Accuracy; TEST = Test Accuracy}
\label{tab:hyperparam-results-001}
\begin{tabular}{|p{3em}|p{3em}|p{3em}|p{3em}|p{3em}|}
\hline
\multirow{3}{*}{HL}& \multicolumn{4}{|c|}{LR = 0.001}\\
\cline{2-5}
& \multicolumn{2}{|c|}{GraphSAGE} & \multicolumn{2}{|c|}{GCN}\\
\cline{2-5}
& VAL & TEST & VAL & TEST \\
\hline
2 & 22.9 & - & 22.7 & -\\
8 & 72.8 & - & 61.3 & -\\
16 & 84.7 & - & 72.4 & -\\
32 & \textbf{85.8} & \textbf{86.67} & \textbf{84} & \textbf{82.6} \\
\hline
\end{tabular}
\end{table}

\begin{figure}[!h]
    \centering
    \begin{subfigure}[b]{0.48\textwidth}
        \centering
        \includegraphics[width=\textwidth]{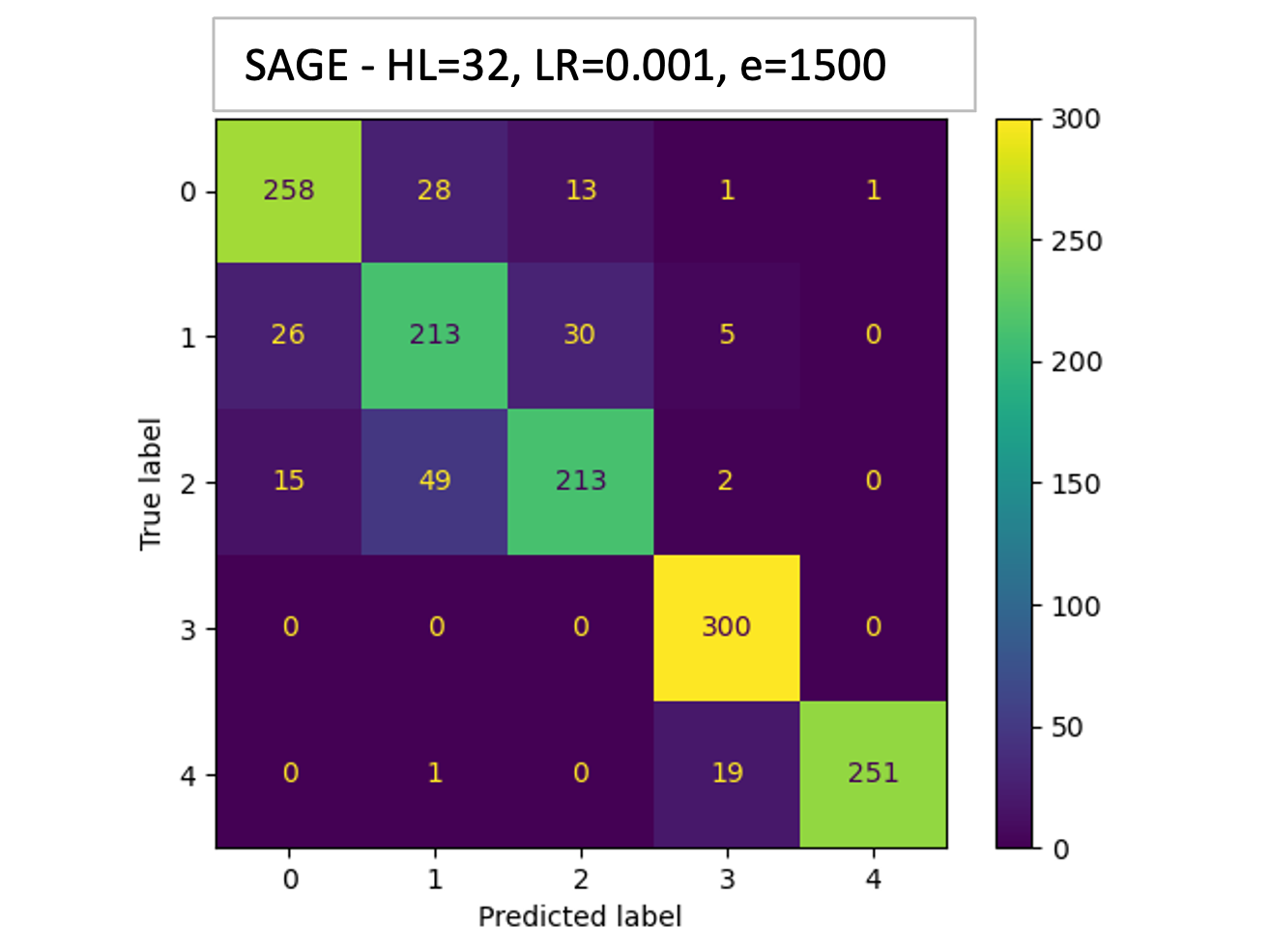}
        \caption{}
        \label{fig:sage-32-001}
    \end{subfigure}
    \hfill
    \begin{subfigure}[b]{0.48\textwidth}
        \centering
        \includegraphics[width=\textwidth]{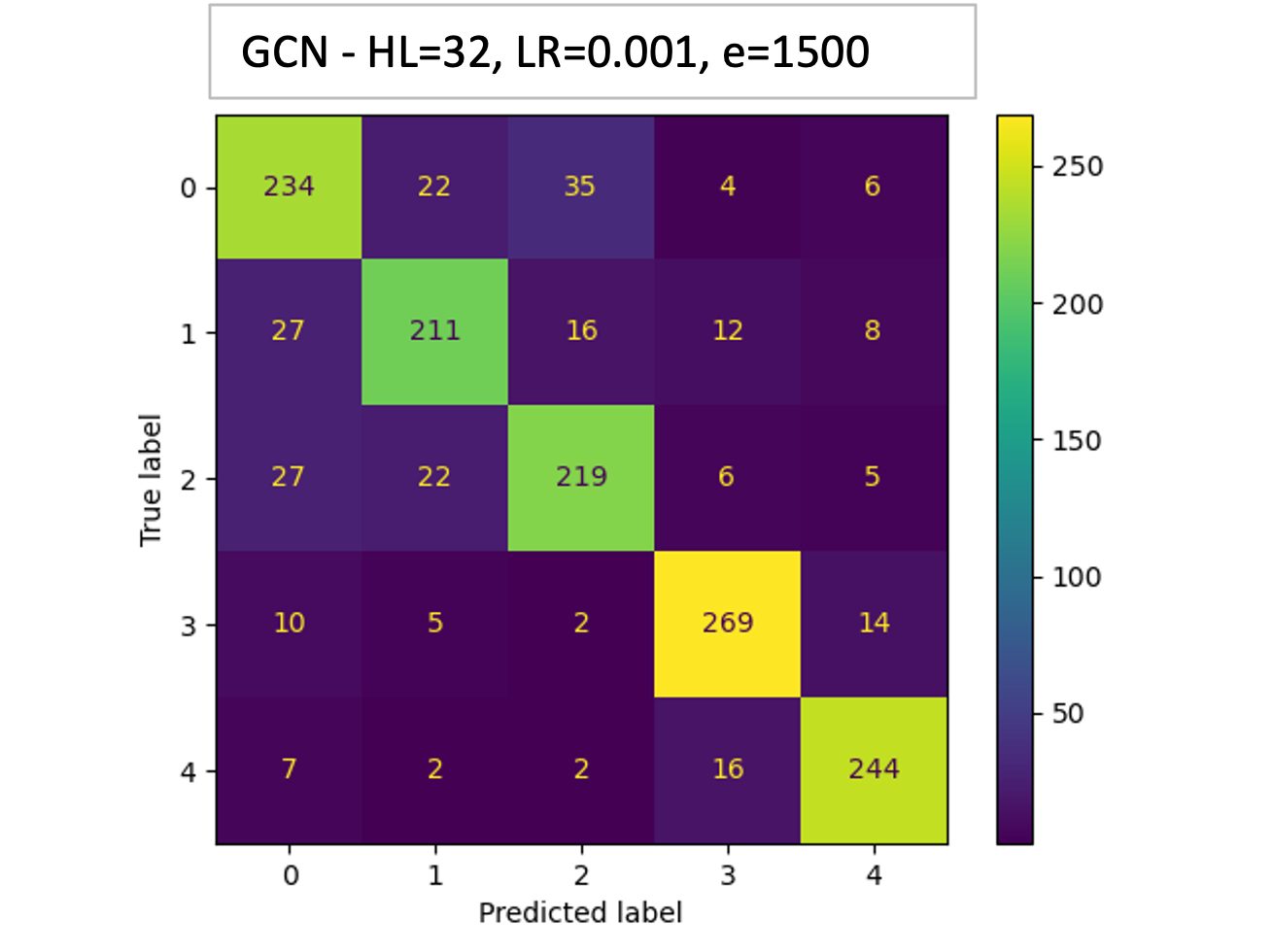}
        \caption{}
        \label{fig:gcn-32-001}
    \end{subfigure}
    \caption{Confusion matrices for (a) GraphSAGE model and (b) GCN model that yielded the highest test accuracy.}
\end{figure}

The confusion matrices for the best models on the test set are shown in Figures \ref{fig:sage-32-001} and \ref{fig:gcn-32-001}.

Another set of experiments of a similar nature, shown in Table \ref{tab:hyperparam-results-01}, were conducted keeping the learning rate constant at 0.01 over 1000 epochs with the same set of hidden layers hyperparameter values. It was observed that the highest validation accuracy for GraphSAGE was 91.5\% with 16 hidden layers and for GCN was 84\% with 32 hidden layers. The best models in these experiments were then evaluated on the test set, giving 91.02\% test accuracy by GraphSAGE over 74.88\% test accuracy by GCN. 

\begin{table}[H]
\centering
\caption{Performance comparison of GraphSAGE and GCN. The test accuracy is only computed for models with the best validation set accuracy. HL = Hidden Layers; LR = Learning Rate; VAL = Validation Accuracy; TEST = Test Accuracy}
\label{tab:hyperparam-results-01}
\begin{tabular}{|p{3em}|p{3em}|p{3em}|p{3em}|p{3em}|}
\hline
\multirow{3}{*}{HL}& \multicolumn{4}{|c|}{LR = 0.01}\\
\cline{2-5}
& \multicolumn{2}{|c|}{GraphSAGE} & \multicolumn{2}{|c|}{GCN}\\
\cline{2-5}
& VAL & TEST & VAL & TEST \\
\hline
2 & 21.5 & - & 21.5 & -\\
8 & 37.9 & - & 60.7 & -\\
16 & \textbf{91.5} & \textbf{91.02} & 81.1 & - \\
32 & 83.1 & - & \textbf{84} & \textbf{74.88} \\
\hline
\end{tabular}
\end{table}

\begin{figure}[!h]
    \centering
    \begin{subfigure}[b]{0.48\textwidth}
        \centering
        \includegraphics[width=\textwidth]{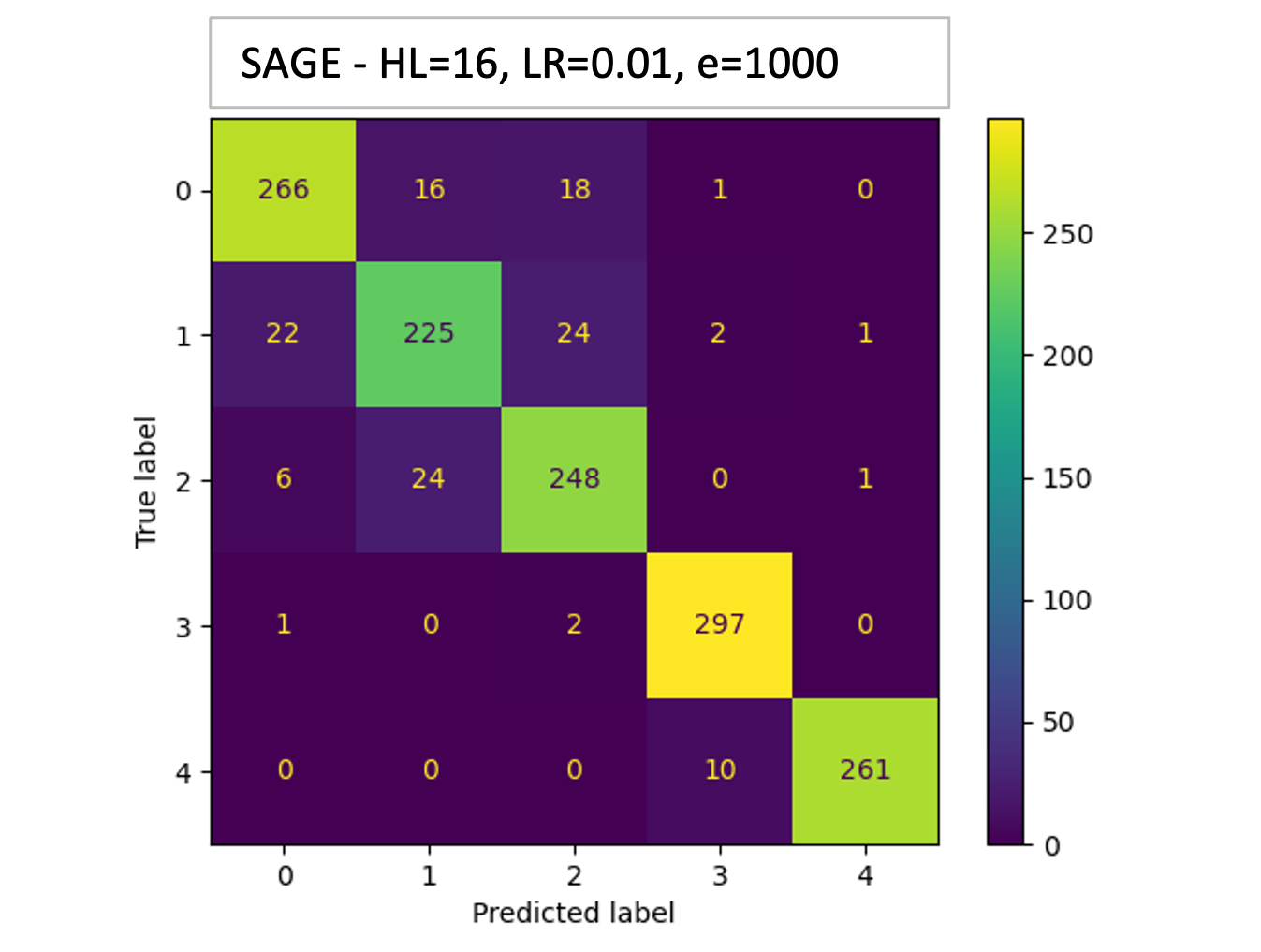}
        \caption{}
        \label{fig:sage-16-01}
    \end{subfigure}
    \hfill
    \begin{subfigure}[b]{0.48\textwidth}
        \centering
        \includegraphics[width=\textwidth]{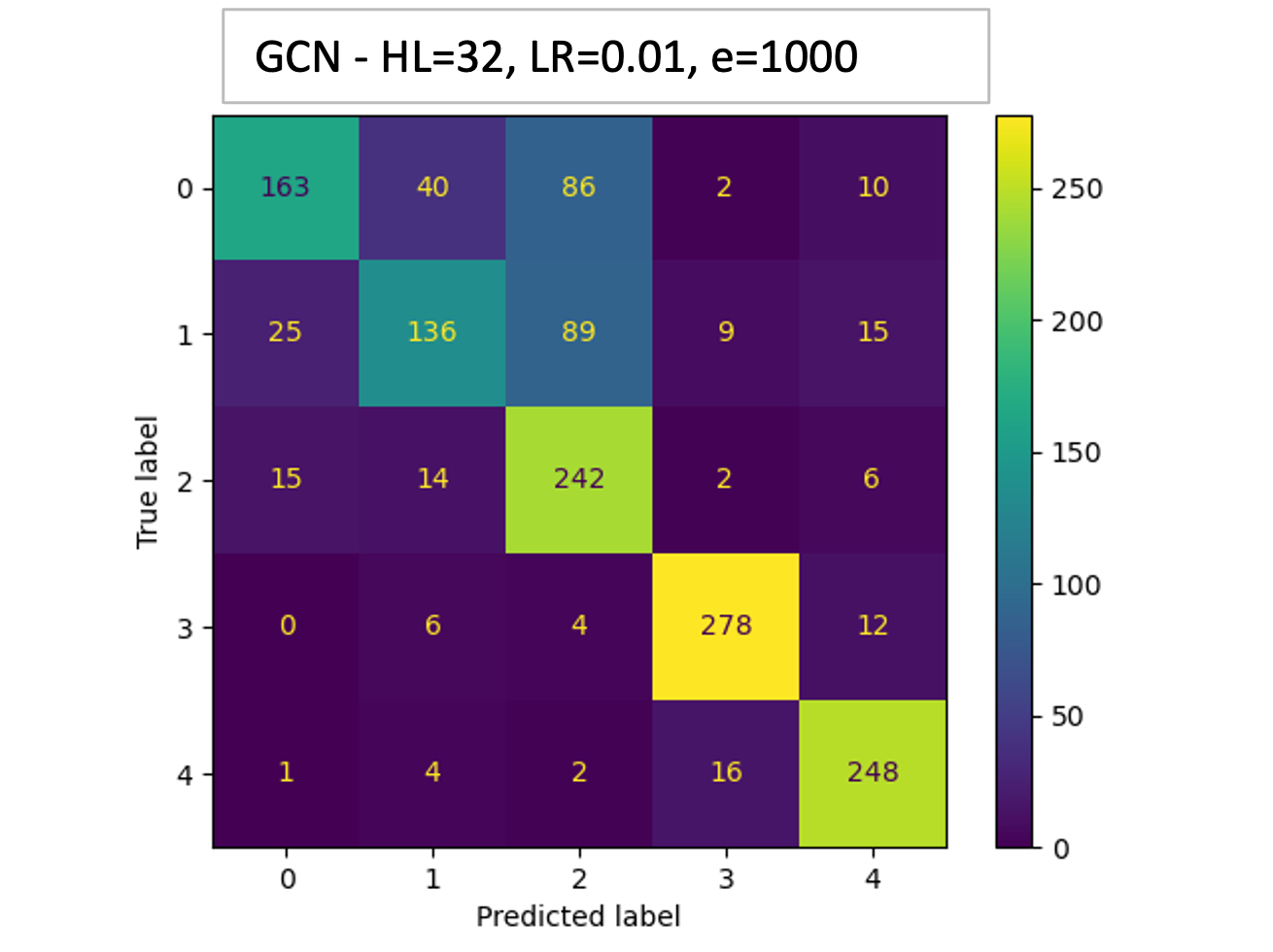}
        \caption{}
        \label{fig:gcn-32-01}
    \end{subfigure}
    \caption{Confusion matrices for (a) GraphSAGE model and (b) GCN model that yielded the highest test accuracy.}
\end{figure}

The confusion matrices for the best models on the test set are shown in Figures \ref{fig:sage-16-01} and \ref{fig:gcn-32-01}. GraphSAGE model with 16 hidden layers and a learning rate of 0.01 gave the highest test accuracy of 91.02\% across all the experiments conducted.

\section{Conclusion}

This work shows that representing genomic data as knowledge graphs allows vast and diverse information to be integrated from various sources. Modeling entities as nodes and relationships as edges provides an ideal framework for integrating and organizing diverse information. We first described the data collection pipeline followed by the usage of the SnpEff tool to obtain additional annotations and SPARQLing Genomics tool to convert the annotations to an RDF format. An ontology to collate information gathered from different sources is presented. Using this ontology, we described how the knowledge graph is created. This knowledge graph contains RNA sequencing information at the variant level from COVID-19 patients from different regions such as lung, blood, etc. Lastly, we presented a case study to demonstrate the usage of the created knowledge graph for a node classification task using the Deep Graph Library. As part of this case study, we also described various experiments that were performed, the corresponding results, and discussed key observations. As part of our future work, we aim to expand the knowledge graph and explore more avenues to use the same to aid researchers working in this domain.

\section{Acknowledgments}

This work was supported by the National Science Foundation under Grant Nos. 2201583 and 2034247.









\bibliographystyle{unsrtnat}
\bibliography{references}  






\end{document}